\documentclass[sigconf]{acmart}
\usepackage{multirow}
\usepackage{tabularx} 
\usepackage{algorithm}
\usepackage{algpseudocode}
\usepackage{float}
\usepackage{enumitem}
\usepackage{svg}

\graphicspath{ {./figures/} }

\AtBeginDocument{%
  \providecommand\BibTeX{{%
    \normalfont B\kern-0.5em{\scshape i\kern-0.25em b}\kern-0.8em\TeX}}}

\copyrightyear{2023}
\acmYear{2023}
\setcopyright{rightsretained}
\acmConference[KDD '23]{Proceedings of the 29th ACM SIGKDD Conference on
Knowledge Discovery and Data Mining}{August 6--10, 2023}{Long Beach, CA,
USA}
\acmBooktitle{Proceedings of the 29th ACM SIGKDD Conference on Knowledge
Discovery and Data Mining (KDD '23), August 6--10, 2023, Long Beach, CA, USA}
\acmDOI{10.1145/3580305.3599897}
\acmISBN{979-8-4007-0103-0/23/08}

\begin{document}

\title{Revisiting Neural Retrieval on Accelerators}

\author{Jiaqi Zhai}
\email{jiaqiz@meta.com}
\orcid{0009-0004-7279-3318}
\affiliation{%
  \institution{Meta Platforms, Inc.}
  \city{Menlo Park}
  \state{CA}
  \country{USA}
  \postcode{94025}
}

\author{Zhaojie Gong}
\email{zhaojieg@meta.com}
\orcid{0009-0004-1761-7530}
\affiliation{%
  \institution{Meta Platforms, Inc.}
  \city{Menlo Park}
  \state{CA}
  \country{USA}
  \postcode{94025}
}

\author{Yueming Wang}
\email{yuemingw@meta.com}
\orcid{0009-0003-3581-8910}
\affiliation{%
  \institution{Meta Platforms, Inc.}
  \city{Menlo Park}
  \state{CA}
  \country{USA}
  \postcode{94025}
}

\author{Xiao Sun}
\email{sunx@meta.com}
\orcid{0000-0002-8385-7020}
\affiliation{%
  \institution{Meta Platforms, Inc.}
  \city{Menlo Park}
  \state{CA}
  \country{USA}
  \postcode{94025}
}

\author{Zheng Yan}
\email{zyan@meta.com}
\orcid{0000-0002-9754-9264}
\affiliation{%
  \institution{Meta Platforms, Inc.}
  \city{Menlo Park}
  \state{CA}
  \country{USA}
  \postcode{94025}
}

\author{Fu Li}
\email{leaf123@meta.com}
\orcid{0009-0004-1141-0870}
\affiliation{%
  \institution{Meta Platforms, Inc.}
  \city{Menlo Park}
  \state{CA}
  \country{USA}
  \postcode{94025}
}

\author{Xing Liu}
\email{xingl@meta.com}
\orcid{0009-0005-0172-8698}
\affiliation{%
  \institution{Meta Platforms, Inc.}
  \city{Menlo Park}
  \state{CA}
  \country{USA}
  \postcode{94025}
}


\begin{abstract}
  Retrieval finds a small number of relevant candidates from a large corpus for information retrieval and recommendation applications. A key component of retrieval is to model (user, item) similarity, which is commonly represented as the dot product of two learned embeddings. This formulation permits efficient inference, commonly known as Maximum Inner Product Search (MIPS).
  Despite its popularity, dot products cannot capture complex user-item interactions, which are multifaceted and likely high rank.
  We hence examine non-dot-product retrieval settings on accelerators, and propose \textit{mixture of logits} (MoL), which models (user, item) similarity as an adaptive composition of elementary similarity functions. This new formulation is expressive, capable of modeling high rank (user, item) interactions, and further generalizes to the long tail. When combined with a hierarchical retrieval strategy, \textit{h-indexer}, we are able to scale up MoL to 100M corpus on a single GPU with latency comparable to MIPS baselines.
  On public datasets, our approach leads to uplifts of up to 77.3\% in hit rate (HR).
  Experiments on a large recommendation surface at Meta showed strong metric gains and reduced popularity bias, validating the proposed approach's performance and improved generalization.

\end{abstract}


\maketitle

\section{Introduction}

Retrieval, also known as candidate generation, is the process of selecting a small number, typically 10-1000, of relevant candidates from a large corpus consisting of millions to billions of items. Retrieval plays a crucial role in large scale information retrieval and recommendation systems as the first stage of the funnel. 
Retrieval must capture (user, item) similarity well by being highly contextualized and personalized; for instance, an individual may want to read important economic news on weekday mornings, but watch lighthearted comedies in the evenings.
\begin{figure*}[t!]
  \centering
  \includegraphics[width=0.95\linewidth]{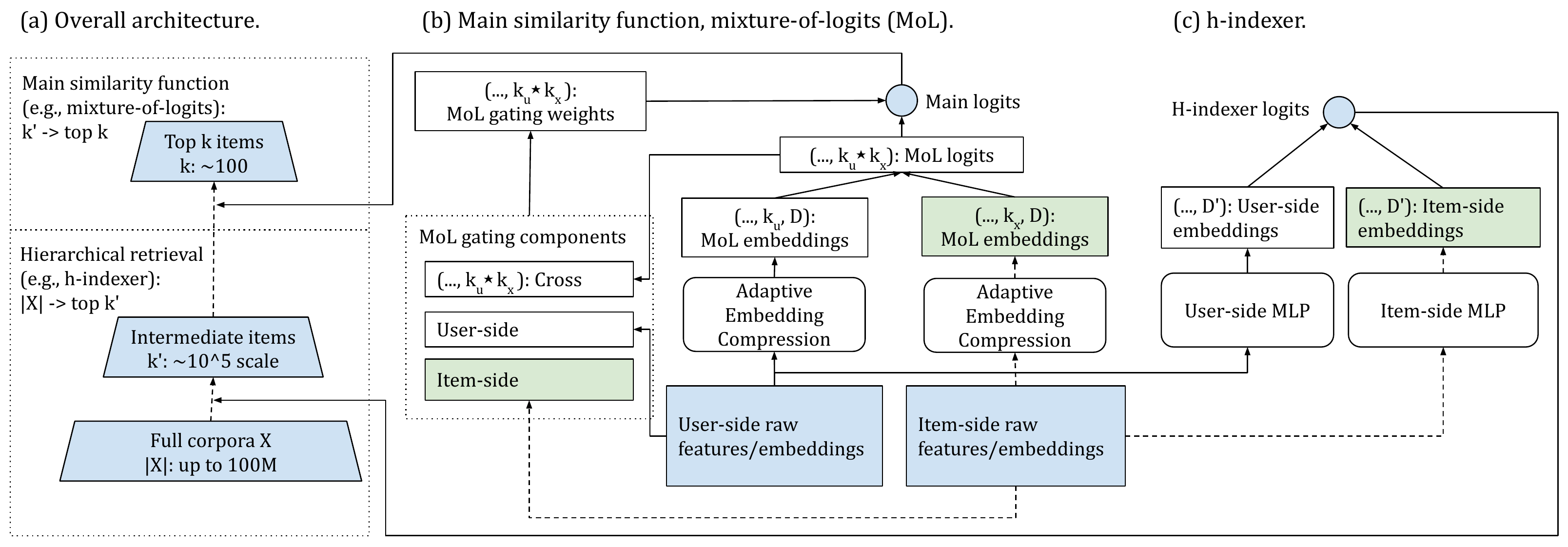} 
  \vspace{-0.9em}
  \caption{Overview of the proposed architecture. Green boxes indicate tensors that are cachable for inference.}
  \label{fig:main-architecture-and-inference-flow}
\end{figure*}

In the deep learning era, (user, item) similarity in retrieval is commonly formulated as the dot product of two embeddings, each parameterized by a deep neural network. This setting is widely adopted throughout industry (e.g., Google~\cite{ytdnn_goog_recsys16,mixed_negative_sampling_goog_www20}, Meta~\cite{ebr_fb_kdd20}, Alibaba~\cite{sdm_baba_cikm19,cl4dcg_baba_kdd21}, etc.). The dot product setting empirically works well for millions of items~\cite{ytdnn_goog_recsys16}, and is computationally efficient; for instance, the inference problem in this setup can be formulated as Maximum Inner Product Search (MIPS). MIPS has been extensively studied, and permits various optimizations such as locality sensitive hashing and product quantization~\cite{alsh_ping_neurips2014,quant_mips_kumar_aistats16}.



\begin{table}[b]
  \vspace{-0.9em}
  \caption{Rank analysis of $\ln p(x|u)$ for various public datasets. }
  \vspace{-0.9em}
  \label{tbl:dataset_rank_new}
  \begin{tabular}{crrrrr}
    \toprule
                &               &              & \multicolumn{3}{c}{Explained variance (SVD)} \\ 
                & \# of users   & \# of items  & $d=64$  & $d=256$ & $d=1024$ \\ 
        \midrule
         ML-1M  &   6,040       &  3,649       & 0.1744 & 0.4126 & 0.8221 \\ 
         ML-20M & 138,493       & 24,186       & 0.1373 & 0.3304 & 0.6452 \\ 
         Beauty &  40,226       & 54,243       & 0.0496 & 0.1212 & 0.2654 \\
         Games  &  29,341       & 23,310       & 0.1100 & 0.2355 & 0.4609 \\ 
         Books  & 694,897       & 674,044      & 0.0336 & 0.0711 & 0.1351 \\ 
  \bottomrule
\end{tabular}
\end{table}

The relationship between users and items in real world, however, demonstrates a significant level of complexity, and may not be approximated well by dot products. There are two evidences. First, dot product-based models produce low-rank recommendations. Consider the log probability of showing item $x$ to user $u$ in the inner product setting, $\ln p(x|u) = \langle f_\theta(u), g_\theta(x)\rangle - \mathbb{Z}_{u}$, where $f_\theta(u), g_\theta(x) \in \mathbb{R}^d$. The $\ln p(x|u)$ matrix hence has a rank of at most $d+1$. Commonly used $d$ values are up to 100s-200s~\cite{ytdnn_goog_recsys16,mind_baba_cikm19,sasrec_icdm18}, which is far less than the rank of actual datasets. As shown in Table~\ref{tbl:dataset_rank_new}, even for small benchmark datasets, dot-product based 256-d models can only explain up to 7.11\%-41.26\% of the variance.  Second, highly expressive neural networks (e.g., high order feature interactions~\cite{wdl_goog_dlrs16,deepfm_ijcai17,afm_ijcai17,dcn_adkdd17}; sequential encoders such as RNNs and Attention~\cite{crarec_neurips16,din_baba_kdd18,sine_baba_wsdm21}) have been the go-to choice for ranking models. A more expressive similarity function, therefore, would better optimize metrics such as HR@10 for the head parts of the corpus.

Despite the above, leveraging expressive similarity functions in retrieval has remained an open ended problem due to (a) the difficulty of training models that generalize, (b) their high inference costs. It's not sufficient to port an arbitrary neural architecture to retrieval; Rendel et al.~\cite{ncf_mf_goog_recsys20} have recently shown that with properly tuned hyper parameters, a learned dot product can substantially outperform multilayer perceptron (MLP)-based learned similarity functions. It's also hard for MLPs to approximate dot products despite MLPs being universal function approximators~\cite{ncf_mf_goog_recsys20,universal_nn_approximator_nn89}.  Computational cost is another major blocker; modern information retrieval applications may need to handle corpus with up to billions of items~\cite{alibaba_scale_kdd18,pixie_pins_www18}. We lack efficient alternatives to MIPS at this scale. One line of work~\cite{tdm_baba_kdd18,jtm_baba_neurips19,otm_baba_icml20} proposes to learn explicit tree structures for the items and reduce top-$k$ similarity to beam search at inference time, which enabled the first successful non-MIPS setup in production. The applications of these approaches have been limited: first, it is generally harder to learn explicit structures --- for instance, prior knowledge are often needed to properly initialize the tree~\cite{tdm_baba_kdd18,jtm_baba_neurips19}, but the tree then needs to change dynamically as the corpus evolves; second, when binary tree structures are used~\cite{tdm_baba_kdd18,jtm_baba_neurips19}, inference latency can be high as beam search requires 22 non-parallelizable steps just to traverse a corpus with 4M items.

To address the above problems, we propose modeling techniques and infra optimizations that enable complex interactions of (user, item) pairs to be modeled at the retrieval stage while supporting efficient inference. We consider the accelerator setting in particular, given that the enablement and proliferation of accelerators like GPUs and TPUs~\cite{tpu-paper} can significantly mitigate the training and serving cost of complex similarity functions. 

\textbf{Our first contribution} is to propose a specific class of high rank similarity functions, \textit{mixture of logits} (MoL), in Section~\ref{sec:mol-algorithm}. MoL is designed with efficient training and inference on accelerators in mind. It not only significantly outperforms dot-product based baselines in hit rate, but also better generalizes to torso and long tail where there are limited training examples by utilizing gating designs and component-level hypersphere embeddings.

\begin{figure}[b]
  \centering
  \vspace{-1.2em}
  \includegraphics[width=0.85\linewidth]{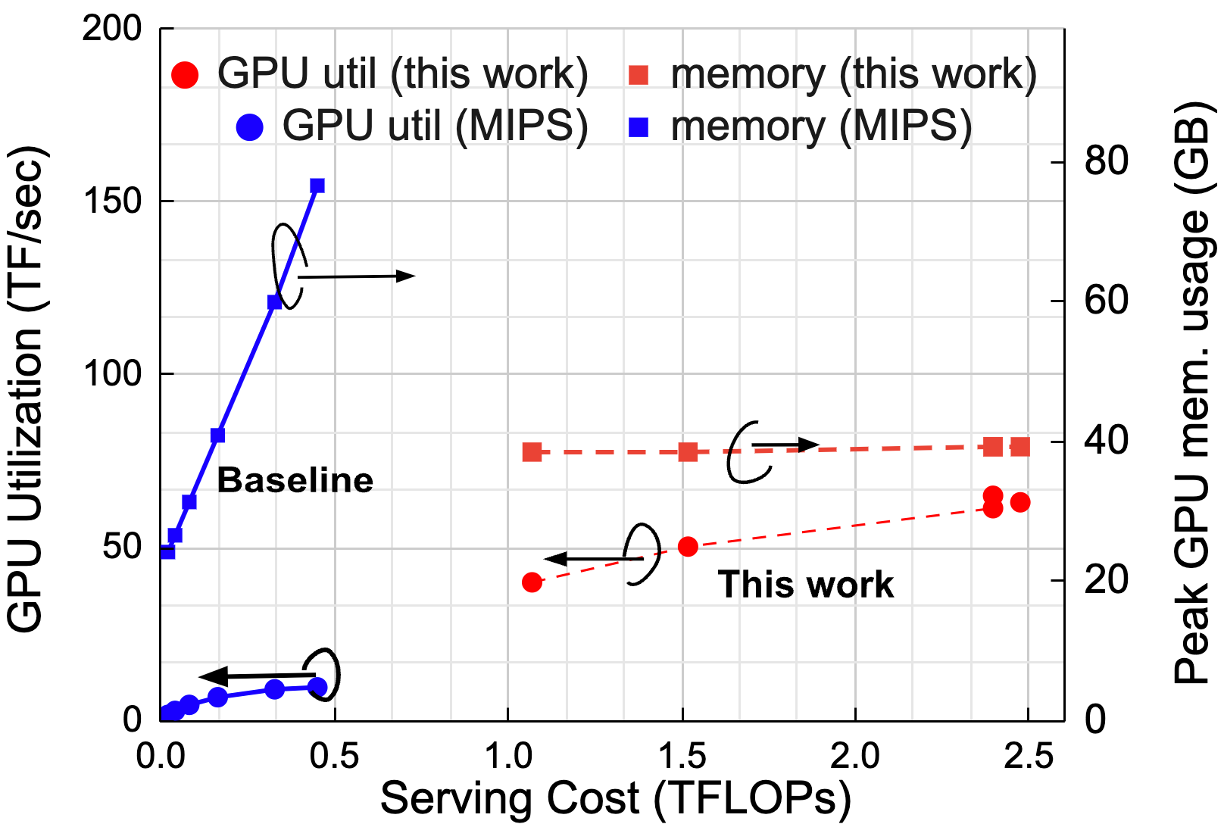}
  \vspace{-1em}
  \caption{Infra efficiency in production: GPU utilization and peak memory scaling with serving FLOPs.}
  \label{fig:gpu-util-memory-scaling}
\end{figure}

\textbf{Our second contribution} is to enable efficient training and inference with expressive similarity functions (e.g., MoL), on modern accelerators (Section~\ref{sec:infra-optimizations}). We propose a hierarchical retrieval strategy, and in particular, an accelerator-friendly algorithm for very large corpus called \textit{h-indexer}. Combined with other optimizations like quantization and kernel fusions, h-indexer enables us to scale up retrieval to hundreds of millions of items with latency comparable to MIPS setups. It's worth remarking that, as highlighted in Figure~\ref{fig:gpu-util-memory-scaling}, the serving cost, e.g., latency and throughput --- on GPUs in this work --- does \textit{not} necessarily increase with compute thanks to the significantly higher GPU utilization of our model. 
Our hierarchical design, unlike dot-product, can also harness more compute with constant GPU memory budget, which is critical given limited memory capacity in production.

\textbf{Our final contribution} is to conduct extensive experiments on public and industrial datasets (Section~\ref{sec:experiments}) with MoL and h-indexer. Our proposed method achieves up to $77.3\%$ gain in HR on standard benchmark datasets like MovieLens and Amazon Reviews, and up to +27.6\% gain at HR@50 over strong baselines on one of the largest recommendation system deployments at Meta. The strong performance of MoL validates our hypothesis that expressive similarity functions are needed to capture dynamics of real-world datasets at retrieval stage. The proposed method also reduces Matthew effect, showing that MoL indeed generalizes well over the entire corpus.  

We finally compare the proposed approach with related works in Section~\ref{sec:related-work} and conclude in Section~\ref{sec:conclusions}.

\section{Overview}
\label{sec:overview}

\subsection{The Retrieval Problem Setting}
\label{sec:background}

In a standard retrieval setup, we are given a user $u$ (or more generally, context $c$) and would like to predict the item $x$'s that is most positively associated with the given user, where $x$'s are selected out of a corpora $\mathbb{X}$. The user $u$ can be characterized by various features, like categorical features 
(e.g., user id, demographics, accounts that the user follows, etc.), float features (e.g., various counters), etc. 

Regardless of the features used, we aim to learn a function $\phi$ parameterized by $\theta$ that encodes similarity between user $u$ and item $x$ as $\phi_\theta(u, x)$. For a given $\phi$, the goal is to learn $\theta$ such that we order $\phi_\theta(u, x)$'s for $x$'s that are positively associated with the user ahead of other $\phi_\theta(u, x')$'s for $x' \in \mathbb{X}$. 

It's useful to formulate the similarity between user $u$ and item $x$ as a probability distribution $p(x|u)$~\cite{ytdnn_goog_recsys16,cl4dcg_baba_kdd21}. We do so by considering $\phi_\theta(x, u)$ as unnormalized logits and pass them through softmax: 
\begin{equation}
p(x|u) = \frac{e^{\phi_\theta(x, u)}}{\sum_{x' \in \mathbb{X}} e^{\phi_\theta(x', u)}} \label{eq:pxu}
\end{equation}

Specific to information retrieval and recommendation setting, the size of $\mathbb{X}$ can be very large, potentially in the range of millions to billions for practical problem settings. Dot products (two tower, dual encoder setups, etc.) are hence commonly used. In this setup, we learn user and item representations as two $d$-dimensional embeddings, $f_\theta(u) \in \mathbb{R}^d$ and $g_\theta(x) \in \mathbb{R}^d$. We recommend item $x$ to user $u$ with probability proportional to their inner products, $\phi_\theta(x, u) \sim \langle f_\theta(u), g_\theta(x)\rangle$; recall that we obtain normalized probability distribution with softmax, hence
\begin{equation}
\ln p(x|u) =  \langle f_\theta(u), g_\theta(x)\rangle - Z_u 
\label{eq:lnp_partition}
\end{equation}
where $Z_u$ is the partition function.

\subsection{Architecture}
\label{sec:architecture}
Figure~\ref{fig:main-architecture-and-inference-flow} shows our main architecture used in the rest of the paper. We highlight the main components of this architecture below.

\textbf{Overall flow}. Our design decomposes retrieval into multiple stages. These stages run in a cascading fashion to produce the final top $k$ candidates. One example is illustrated in Figure~\ref{fig:main-architecture-and-inference-flow}(a), where an accelerator friendly algorithm, \textit{h-indexer}, is used to find a large number of candidates ($k'=10^5$) out of a 100M corpus, and then a complex similarity function (e.g., \textit{mixture-of-logits}) is used to find the final top $k$  (e.g., $100$) candidates. We show in Section~\ref{sec:exp-hierarchical-retrieval-eval} that this design significantly improves throughput without degrading recall for suitably chosen values of $k'$.

\textbf{Main similarity function}. The hierarchical retrieval design enables complex neural networks to be used when modeling (user, item) similarities. In Section~\ref{sec:mol-algorithm}, we discuss one such instance of similarity functions, \textit{mixture of logits}, that significantly outperforms dot products. When designing this architecture, we aggressively make intermedidate tensors available for caching (green boxes in Fig.~\ref{fig:main-architecture-and-inference-flow}). For instance, we can cache item-side gating weights and combine them cheaply with non-cachable weights at inference time.

\textbf{h-indexer}. We found that a simple, but highly optimized dot product combined with specialized top-$k$ algorithm works well for up to 100M corpus. This stage is co-trained with the main similarity function. We discuss this design in details in Section~\ref{sec:infra-alg-prefilter}.

\section{Mixture of Logits: An Accelerator-Aware Model Design}
\label{sec:mol-algorithm}

In this section, we propose a new high rank similarity function, \textit{mixture of logits} (MoL). MoL is designed for neural retrieval settings on accelerators. The basic idea in MoL is to parameterize $\phi_\theta(u, x)$ 
as an \textit{adaptive} mixture of  more elementary logits,

\begin{equation}
\phi_{MoL}(x, u) = \sum_k \pi_{k, \theta}(x, u) \delta_{k, \theta}(x, u) \label{eq:generalized_mol}
\end{equation}
where $\pi_{k, \theta}(x, u)$ are adaptive gating weights and $\delta_{k, \theta}(x, u)$ are elementary logits.

MoL achieves high rank in two ways. First, since $\delta_{k, \theta}(x, u)$ in Equation~\ref{eq:generalized_mol} can be parameterized by any form of neural networks, $\phi_\theta(x, u)$ has arbitrarily high rank. 
Second, as $\pi_{k, \theta}(x, u)$ takes $x$ and $u$ as input, we can create a high rank matrix $\ln p(x|u)$ by combining low rank similarity functions, e.g., dot products, in a cost-efficient fashion. For instance, using $k$ groups of dot products (of potentially different embedding dimensions), we can derive 

\begin{equation}
\phi_{MoL_{dot\ products}}(x, u) = \sum_k \pi_{k, \theta}(x, u) \langle f_{k, \theta}(u), g_{k, \theta}(x)\rangle \label{eq:mol_dot_products_def}
\end{equation}

Although the elementary logits $\langle f_{k, \theta}(u), g_{k, \theta}(x)\rangle$ themselves have ranks of at most $d$ (inner products of $d$-dimensional embeddings), Equation~\ref{eq:mol_dot_products_def} 
 can define an arbitrarily high rank probability distribution, as long as $\pi_{k, \theta}(x, u)$ depends on \textit{both} item $x$ and user $u$. Otherwise, suppose $\pi_{k, \theta}(x, u)$ only depends on $u$, we have
\begin{align*}
\sum_k \pi_{k, \theta}(u) \left\langle f_{k, \theta}(u), g_{k, \theta}(x)\right\rangle =  \sum_k  \left\langle \pi_{k, \theta}(u) f_{k, \theta}(u), g_{k, \theta}(x)\right\rangle
\addtocounter{equation}{1}\tag{\theequation}
\label{eq:mol_low_rank_variant}
\end{align*}
which defines $\phi_{MoL}(x, u)$ of rank at most $kd$ when all embeddings are of dimensionality $d$. Similar argument holds on the item side.

Empirically this formulation does enable us to achieve high rank, as shown in Section~\ref{sec:exp-public-experiments}.

\subsection{Embedding Sharing and Adaptive Embedding Compression}
\label{sec:alg-emb-sharing-adaptive-logits}

Implementing Equation~\ref{eq:mol_dot_products_def} directly requires $k$ matrix multiplication (\textit{mm}) / batched matrix multiplication (\textit{bmm}) operations. 
We can accelerate training and inference as follows: (a) make embedding dimensions used across different mixture components the same, and (b) reuse embeddings across these groups and enable $k$ to be different on the user and on the item side. Specifically, we consider $k_u$ embeddings of dimensionality $d$ on the user side and $k_x$ embeddings of the same dimension on the item side. We can then take dot products between all $k_u$-by-$k_x$ pairs of embeddings:

\begin{equation}
\phi_{MoL_{dot\ products'}}(x, u) = \sum_{k_u, k_x} \pi_{k_u, k_x, \theta}(x, u) \langle f_{k_u, \theta}(u), g_{k_x, \theta}(x)\rangle \label{eq:mol_def_cross_dot_products}
\end{equation}

These logits in Equation~\ref{eq:mol_def_cross_dot_products} can be computed with one $bmm$ for inference, and one $mm$ for training as will be shown in Section~\ref{sec:alg-gating-decomposition}.

\subsubsection{Adaptive Logits Selection} The $k_u$ user-side and $k_x$ item-side embeddings in Equation~\ref{eq:mol_def_cross_dot_products} can be interpreted as (transformations) of user-side and item-side features (e.g.,~\cite{wdl_goog_dlrs16,deepfm_ijcai17}). The drawback is that there could be hundreds to thousands of such features in production settings, which makes computing $k_u \times k_x$ logits prohibitively expensive. We control computational cost as follows:
\begin{itemize}[leftmargin=*]
    \item \textit{user-side / context-side}. 
    We reduce computational cost by compressing the feature embeddings \textit{adaptively} (`Adaptive Embedding Compression' in Figure~\ref{fig:main-architecture-and-inference-flow}(b)). We transform the $k'_{u}$ embeddings $v'_i$'s into $k_{u}$ embeddings $v_{i}$'s of the same dimensionality:
    \begin{equation}
        v_{i} = \sum_{0 \leq j < k'_u} w_{j,i} v'_{j}. \label{eq:adaptive-logit-selection-user-side}
    \end{equation}
    \item \textit{item-side}. Besides computational challenge, accelerator memory limits how many embeddings we can store. On a 16GB accelerator and with a 10M corpora where $d=128$, we can only cache 6 item-side embeddings with FP16. Besides using Equation~\ref{eq:adaptive-logit-selection-user-side} for compression, we can generate some item-side embeddings on the fly with learned MLPs, which trades off compute for memory. 
\end{itemize}

\subsection{Shared Negatives and Gating Function Decomposition}
\label{sec:alg-gating-decomposition}

So far we have omitted discussions about how $\pi_{k, \theta}(x, u)$ is defined. Modern neural retrieval models are commonly trained with sampled softmax losses~\cite{ytdnn_goog_recsys16,sdm_baba_cikm19,mixed_negative_sampling_goog_www20,sine_baba_wsdm21}, and utilizing more negatives generally improves model quality~\cite{mixed_negative_sampling_goog_www20}. 
Even if we share the negatives for samples in a mini-batch, the large number of negatives can still make popular gating/attention architectures which are originally designed for ranking models prohibitively expensive in terms of computation and memory. For example, in AttentionFM~\cite{afm_ijcai17}, the attention weights for logits are computed as 
$
\alpha_{ux} = \textrm{MLP}(v_u \odot v_x)
$ where $\odot$ is element-wise dot product.
With the extra item dimension in large batch negative training, the computational cost for attention weights becomes $O(BX(k_uk_x)^2D_f)$ for one linear layer, where $B$ is mini batch size, $X$ is number of negatives, and $D_f$ is the feature embedding dimension. This requires $35184.4$ GFLOPs and 274.9GB HBM for configurations discussed in Table~\ref{tbl:gating_decomposition} ($D_f = 256$).


Therefore, in addition to sharing negatives across users in a mini-batch as done in standard negative sampling~\cite{ytdnn_goog_recsys16, mixed_negative_sampling_goog_www20,cbns_huawei_sigir21} and removing element-wise dot products, we further propose to rewrite the gating function with decomposition to reduce the computational cost of materializing and performing matrix multiplications over large $(B, X, D)$ tensors.
This is done by parameterizing $\pi_{\theta}(x, u)$ as 
\begin{equation}
    \pi_{\theta}(x, u) = \sigma_\theta\left(\pi^X_{\theta}(x), \pi^U_{\theta}(u), \pi^{XU}_{\theta}(x|u)\right)
\end{equation}
where $\pi^X_{\theta}(x)$ is a DNN that takes item-side feature of dimensionality $D^{X}$ as input, $\pi^U_{\theta}(u)$ is a DNN that takes user-side feature of dimensionality $D^{U}$ as input, and $\pi^{XU}_{\theta}(x|u)$ takes cross features of dimensionality $D^{XU}$ (e.g. $k_x \cdot k_u$) as input, and $\sigma_\theta(\cdot, \cdot, \cdot)$ is some nonlinearity (e.g., softmax after linear combination of inputs). Assuming we use two-layer MLPs with $K$ (or $K^U$, $K^X$, $K^{XU}$) as hidden dim and $L$ as output dim, the original computational cost of $\pi_{\theta}(x, u)$ is $O(BXK(D+L))$, which is reduced to $O(BK(D^U+L)+XK(D^X+L)+BXK(D^{XU}+L))$ with gating function decomposition. With decomposition of user, item, user-item tensors, we are able to reduce the computational cost of the most expensive part (with $D^{XU} \ll D$) while maintaining model quality. We show the benefit of this approach for practical settings in Table~\ref{tbl:gating_decomposition}. 

%

\begin{table}
  \caption{Effect of $\pi_{\theta}(x, u)$ decomposition. $B = 2048$, $X = 4096$, $D = 1024$ ($D^U=768$, $D^X=128$, $D^{XU}=128$), $K=256$, $L=128$.}
  \vspace{-0.9em}
  \label{tbl:gating_decomposition}
  \begin{tabular}{cll}
    \toprule
                                      & GFLOPs     & HBM (fp32) \\
        \midrule
         w/o gating decomposition     & 2473.9     &   44GB     \\
         w/ gating\ decomposition     & \textbf{1101.0 (-55.5\%)}    &   \textbf{16GB (-63.6\%)}    \\
  \bottomrule
\end{tabular}
\end{table}

Finally, we regularize $\pi_{\theta}(x | u)$ by forcing it to take the form of a probability distribution per (user, item) pair, i.e., $\sum_k \pi_{k, \theta}(x | u) = 1$. We found that applying dropout on top of this distribution to be helpful, which encourages more balanced mixture component utilization similar to Mixture-of-Experts~\cite{moe_nc91,hmoe_nc93}. 

\subsection{Component-level Hypersphere Embeddings}

The large number of component-level embeddings used in Equation~\ref{eq:mol_dot_products_def} and~\ref{eq:mol_def_cross_dot_products} could introduce two problems, preventing generalization of the model to the torso and long tail in retrieval setting:
\begin{itemize}
    \item The gating function $\pi_{\theta}(x, u)$ can make component-level embeddings undertrained (esp. in online training settings), which in turn introduces training instability over time. 
    \item They may exacerbate overfitting or require significantly more negatives (more compute) to properly fit the model.
\end{itemize}

We propose L2 normalizing the component embeddings as a simple and effective regularization to fix the problem. The intuition is as follows. L2 normalization forces all MoL component embeddings in Equation~\ref{eq:mol_def_cross_dot_products} to lay on the surface of a $d$-dimensional hypersphere. Denote the volume of a $d$-dimensional hypersphere by $V(d)$. We have $V(d) \geq \frac{1}{\sqrt{d-1}} V(d-1)$; as $d$ grows to infinity, the majority of the volume of a unit hypersphere lies near its `equator' which is in turn a $(d-1)$ dimensional hypersphere. Further, the majority of a unit hypersphere's volume lies near its surface. This property encourages component-level hypersphere embeddings, even if undertrained, to be near-orthogonal to each other: $\lim_{d \to \inf} \langle f_{k_u}(u), g_{k_x}(x)\rangle / || f_{k_u}(u)||_2 / || g_{k_x}(x) ||_2 = 0$. In contrast, without normalization, assume the dimensions of each embedding are i.i.d. in [-1, 1], the limit of the above equation goes to infinity. 
Formally, the modified MoL setup can be found in Equation~\ref{eq:mol_def_cross_dot_products_l2}, where the term $\tau$ ($\tau > 1$) rescales post softmax output range back to $(0, 1)$, similar to temperature in L2-normalized softmax.

\begin{equation}
\phi_{MoL_{final}}(x, u) = \sum_{k_u, k_x} \pi_{k_u, k_x, \theta}(x, u) \left\langle \frac{f_{k_u, \theta}(u)}{||f_{k_u, \theta}(u)||_2}, \frac{g_{k_x, \theta}(x)}{||g_{k_x, \theta}(x)||_2}\right\rangle/\tau \label{eq:mol_def_cross_dot_products_l2}
\end{equation}

Besides solving the problem of undertrained component-level embeddings, the other nice property of Equation~\ref{eq:mol_def_cross_dot_products_l2} is that we force MoL logit components to measure angular distances instead of both vector magnitudes and angular distances, which helps avoiding degrading the model to only learn about popular items given impressions on items generally follow a power-law distribution. We present ablation studies in Section~\ref{sec:exp-public-ds-ablation}. 

\subsection{Final MoL Algorithm} Pseudocode for the final mixture-of-logits model using dot products as mixture components 
can be found in Algorithm~\ref{alg:mol-with-prefilter}, which is also illustrated in Figure~\ref{fig:main-architecture-and-inference-flow}(b). In practice, we use simple 2-layer MLPs for $queryWeightFn$, $itemWeightFn$, and $crossWeightFn$ and set their output size to be $k_u \cdot k_x$. We use a simple $combinationFn$, $f(uw, xw, cw) = uw \cdot xw + cw$ followed by SiLU nonlinearity~\cite{silu_nn2017}.

\begin{algorithm}
\caption{Mixture of Logits (MoL) Algorithm.} 
\label{alg:mol-with-prefilter}
\begin{algorithmic}[1]
\Procedure{MoL}{$data, query, k, k', k_u, k_x, d, \tau$}
    \State $userEmbs \gets \textsc{l2Norm}(userEmbProj(query))$ 
    \State $itemEmbs \gets []$  \Comment{Of shape ($k_x, d$) $\times$ $k'$}
    \For{$x \in data$} \Comment{Cachable}
        \State $itemEmbs.append(\textsc{l2Norm}(itemEmbProj(x)))$
    \EndFor
    \State $cl \gets mm(userEmbs, itemEmbs.reshape(-1, d).t()) / \tau$
    \State $cl \gets cl.view(k_u, k', k_x).permute(1, 0, 2).reshape(k', -1)$
    \State $gatingWeights \gets \textsc{decomposedGating}(data, query, cl)$ 
    \State $similarities \gets (gatingWeights \cdot cl).sum(-1)$
    \State \textbf{return} $argsort(similarities)[:k]$
\EndProcedure
\Procedure{decomposedGating}{$data, u, allLogits$}
    \State $uWeights \gets userWeightFn(u)$
    \State $xWeights \gets itemWeightFn(data)$         \Comment{Cachable}
    \State $crossWeights \gets crossWeightFn(allLogits)$  
    \State \textbf{return} \textbf{softmax}$(combineFn(\newline uWeights, xWeights, crossWeights))$   \Comment{Of shape $(k', k_u \cdot k_x)$}
\EndProcedure
\end{algorithmic}
\end{algorithm}

The time complexity for caching item-side computations in MoL is $|\mathbb{X}| \cdot (O(itemEmbProj) + O(itemWeightFn))$, which is negligible on accelerators. Assuming all 2-layer MLPs have a hidden dimension of $d_h$, user embedding input dimension is $d_u$, and all cachable computation are cached, the MoL stage has a computational cost of $O(userEmbProj) + O(k_u \cdot d + k' \cdot k_u \cdot k_x \cdot d) + O(userWeightFn) + k' \cdot O(crossWeightFn) + k' \cdot O(combineFn) + O(k' \cdot k_u \cdot k_x)$. This cost is dominated by the user-item cross parts: $O(k' \cdot k_u \cdot k_x \cdot d + k' \cdot O(crossWeightFn) + k' \cdot O(combineFn)) = O(k' k_u k_x (d + d_h))$.


\section{Hierarchical Retrieval and Infrastructure Optimizations}
\label{sec:infra-optimizations}

We made the following optimizations to enable large-scale MoL based retrieval model training and serving in production.


\subsection{H-indexer Design and Optimization}
\label{sec:infra-alg-prefilter}

In the hierarchical retrieval design (Section~\ref{sec:architecture}), the first stage finds a large number of candidates ($k' \sim 10^5$) with the goal of not missing any candidates that would have originally been selected by the final similarity function in top $k = 100 \sim 1000$. The high value of $k'$ makes blockwise based top-k algorithms such as FAISS~\cite{faiss_tbd21} and TPUKNN~\cite{tpuknn_goog_neurips22} less applicable, where small $k's$ ($\leq 2048$) are preferred due to register size limit or algorithm design. We therefore propose an approximate top-k design for very large $k$s, \textit{h-indexer}, with pseudocode in Algorithm~\ref{alg:prefilter}. 
We discuss optimizations used below.

\subsubsection{dot-product optimization} 
To control the computational cost of dot-product stage, we used a low embedding dimension (64). 
Meanwhile, dot product is memory-bounded and its arithmetic intensity --- a measure indicating the degree of an op being compute-bound over memory-bound~\cite{roofline-arithmetic-intensity} --- only scales with batch size and reversely with input byte-width (byteW),
\begin{equation}
    A.I. = \frac{B\times X\times D'\times 2FLOPs}{(B\times D' + X\times D' + B\times X)\times byteW}\approx \frac{2B}{byteW}
\end{equation}
where corpus size (X) and embedding dimension (D) are far greater than batch size (B). With this observation, we boosted the GPU utilization by batching user requests and applying INT8 quantization. Dot product shows $1.5\times$ speed boost compared to the half precision thanks to INT8. Further, the output of INT8 GEMM is INT32, which can be digested by top-k selection directly without applying scale/bias to de-quantize to FP32 as in the regular practice.

\begin{algorithm}[t]
\caption{h-indexer top-$k$ algorithm for large $k$s ($k'$s). $\lambda$ is the selection ratio for estimating top $k$ similarity threshold.}
\label{alg:prefilter}
\begin{algorithmic}[1]
\Procedure{h-indexer}{$data, query, k, \lambda$}
    \State $randIndices \gets \textsc{permutation}(0..|data|-1)[: \lambda]$
    \State $sampledSimilarities \gets []$
    \For{$x \in data[randIndices]$}
        \State $sampledSimilarities.append(\textsc{dotProduct}(x, query))$
    \EndFor
    \State $t \gets \textsc{nthElement}(sampledSimilarities, k / |data| \cdot \lambda)$
    \State $indices \gets []$
    \For{$(indice, x) \in \textsc{enumerate}(data)$}
        \If {$\textsc{dotProduct}(x, query) > t$}
            \State $indices.append(indice)$
        \EndIf
    \EndFor
    \State \textbf{return} $indices$
\EndProcedure
\end{algorithmic}
\end{algorithm}
\subsubsection{approximate top-k} Finding top-$k'$ in a corpus with $X$ items 
has a time complexity of $\Omega{(X \log k')}$, which is exacerbated by the large $k'$ number. We therefore adopted an approximate top-k approach to randomly sample a small portion of the corpus to estimate the threshold for top $k'$ items, and subsequently finding up to $k'$ samples that meet that threshold. This reduces time complexity to $\Omega{(X + rX \log k')}$ with the sampling ratio $r$ ranges from 0.01 to 0.1. In production settings, this change is ${\sim}2.5\times$ faster than exact top-k and reduces h-indexer latency by $30\%$. 

\subsubsection{index select} We further optimized the index select op after top-k selection in Algorithm~\ref{alg:prefilter}. Fetching and concatenating $10^5$ item embeddings out of a 10M pool is expensive with a throughput of around {0.15 TB/s}. We designed an optimized GPU kernel with $2\times$ throughput by leveraging data locality and efficient caching \cite{fbgemm-index-select}.

\subsection{Op Fusion for Latency and Memory Saving}
We also performed Op fusion to further reduce the serving cost for the MoL stage and achieved 30\% performance improvement. While GEMM ops are highly optimized on GPUs, Armdahl's law dictates that non-GEMM ops would become a bottleneck for the end-to-end latency. The read/write cost of input/output tensors between HBM and GPU's caches is significantly exacerbated by the large item size ($10^5$). Fusing consecutive non-GEMM ops, such as $SiLU(x)=x \cdot \text{Sigmoid}(x)$, and fusing non-GEMM ops as epilogues of GEMM could efficiently reduce R/W induced latency. Op fusion can also reduce memory usage for MoL. For example, the op below requires expansion of input tensors by $B=8$ or $K'=10^5$ for the operands in $\text{SoftMax}$, which can be avoided in a fused kernel.
\begin{equation*}
    Out = [B, K', D] \times \text{SoftMax} ([B, D] + [K', D])
\end{equation*}
We also fused line 7 and 8 in Algorithm~\ref{alg:mol-with-prefilter}. Since the size of $cl$ is much larger than $userEmbs$ and $itemEmbs$, fusing the GEMM op and the permute op can greatly save memory and improve throughput.
\vspace{-0.5em}
\subsection{BFLOAT16 Computation}
Although FP16 training has demonstrated $2\times$ throughput than FP32 training~\cite{nvidia-fp16-paper}, the range of FP16 (1-5-10) is insufficient for the backwards gradients in training. The compensating loss scaling technique discards the overflowed iteration unpredictably and results in poor reproducibility for production. We therefore explored the training in BF16 with the same range as FP32. We further aggressively applied BF16 precision to ops like SoftMax, LayerNorm, and Divison and observed no apparent degradation of the final metrics. It is important to perform these non-GEMM ops in BF16 to avoid the time- and memory-costly conversions between BF16 and FP32. Once the model is trained in BF16, we inference it in the FP16 precision. The $BF16\rightarrow FP16$ conversion introduces no quantization noise as FP16 has more mantissa bits than BF16. Eventually, training throughput is increased by 22\% with BF16 computations.
\vspace{-0.5em}
\subsection{FP8 Quantized Communication} Another innovation we performed in this study is aggressive quantization of All2All traffic to the FP8 format~\cite{hybrid-fp8-paper, fp8-standard}. As a costly bottleneck for recommendation model training, All2All communication that switches between data parallelism and model parallelism is especially expensive for our retrieval model thanks to the complex user embeddings from features and from sequential encoding. FP8 quantization has only 8-bit bitwidth and can handle both forward activations and backward gradients with dynamic scaling. We created a customized FP8 rowwise quantization kernel~\cite{fbgemm-index-select} to quantize all2all tensors before the communication. Since All2All does not require any arithmetic operations, the FP8 rowwise quantization kernel can be executed by A100 GPU that does not even support FP8 arithmetics. The resulting final metrics show no degradation and gained 16\% throughput over 16-bit All2All communication. In combination, quantized training with BF16 computation with FP8 communication improved throughput by 37\%.
\vspace{-0.5em}



\section{Experiments}
\label{sec:experiments}

We analyze the proposed techniques on both public datasets and large scale industrial datasets. Our proposed techniques are agnostic to the exact retrieval model setup, and apply as long as we need to model (user, item) similarities. We hence consider two settings:

\begin{itemize}[leftmargin=*]
    \item \textit{Non-sequential methods.} This is a standard retrieval setup where the goal is to model the probability of recommending a candidate item given a user as in Equation~\ref{eq:pxu}, and is commonly used in various retrieval and recommendations settings~\cite{ytdnn_goog_recsys16,ncf_www17,mind_baba_cikm19,sdm_baba_cikm19,tdm_baba_kdd18,jtm_baba_neurips19,cl4dcg_baba_kdd21,dr_cikm21}. Note that
    while the model setup is not fully sequential,
    sequential encoders 
    are commonly used in these setups to provide features as user-/context-side encoders~\cite{tdm_baba_kdd18,mind_baba_cikm19,sdm_baba_cikm19,jtm_baba_neurips19,cl4dcg_baba_kdd21}.

\item \textit{Sequential methods.} These methods predict next item based on sequential user-item interactions~\cite{sasrec_icdm18,bert4rec_cikm19,ssrlrec_sigir20}. 
    User interactions are modeled as a (higher-order) Markov chain. 
\end{itemize}

We tuned hparams for all non-dot-product setups (MLP, NeuMF~\cite{ncf_www17}, DeepFM~\cite{deepfm_ijcai17}, MoL) so that they use similar compute. We omitted AttentionFM~\cite{afm_ijcai17} since it cannot fit into 80GB HBM for most configurations due to AFM's usage of element-wise products combined with large number of negatives, as discussed in Section~\ref{sec:alg-gating-decomposition}. 

\begin{table}
  \caption{Data statistics (after preprocessing and truncation)}
  \vspace{-0.9em}
  \label{tbl:public_dataset_stats}
  \begin{tabular}{crrrr}
    \toprule
                       & \# of users        & \# of items   & \shortstack{avg. ratings \\per item} & \shortstack{avg. ratings \\ per user} \\
        \midrule
         ML-1M         &   6,040            &   3,649       &  181.59     & 109.70  \\
         ML-20M        & 138,493            &  24,186       &  528.40     &  92.28  \\
         Beauty        &  40,226            &  54,243       &   6.40      &   8.62  \\
         Games         &  29,341            &  23,310       &  11.97      &   9.15  \\
         Books         & 694,897            & 674,044       &  11.51      &  11.61  \\
  \bottomrule
\end{tabular}
\end{table}

\subsection{Public datasets}

We consider MovieLens (1M and 20M variants), Amazon Reviews (Beauty, Games, and Books subsets) for consistency with prior work~\cite{sasrec_icdm18,bert4rec_cikm19,sdm_baba_cikm19,jtm_baba_neurips19,cl4dcg_baba_kdd21}. 
We follow standard preprocessing, including discarding users and items with fewer than 5 related actions, and truncating user behaviorial sequences to 200 (MovieLens) / 50 (Amazon)~\cite{sasrec_icdm18,bert4rec_cikm19}. Statistics for all datasets can be found in Table~\ref{tbl:public_dataset_stats}.

For these datasets, due to the popularity of sequential methods in recent years, we apply our proposed approach on top of a sequential baseline, SASRec~\cite{sasrec_icdm18}. SASRec trains the model to predict the next item in an autoregressive manner. 
SASRec represents one of the best known approaches on public datasets; for instance, on ML-1M, our SASRec implementation achieves $.223$ HR@10 (Table~\ref{tab:public_movielens}), on par with best result ($.224$) from~\cite{seqreceval_recsys21}, and on Amazon Games, our impl's HR@10 is $.0794$, which is within $10\%$ of the best result in~\cite{seqreceval_recsys21}. 


\subsubsection{Evaluation Methodologies} Prior work have reported metrics based on different sampling methodologies; e.g., Kang et al. used 100 randomly sampled negatives~\cite{sasrec_icdm18}, and Sun et al. used 100 popularity-weighted sampled negatives~\cite{bert4rec_cikm19}. For consistency with standard retrieval settings, we report metrics evaluated over the entire corpus, which is also done in~\cite{seqreceval_recsys21}. We omit \textit{h-indexer} given the corpus are small. We focus on HR@10 for MovieLens and HR@200 for Amazon Reviews based on prior work.

\subsubsection{Experiments}
\label{sec:exp-public-experiments}

We consider the baseline setting (binary cross entropy loss on top of dot products, BCE) and also other settings using sampled softmax (SS), including Dot Product, MLP, NeuMF, DeepFM, and MoL. 
We train all variants with Adam~\cite{adam_iclr15}, a learning rate of $0.001$, and a mini batch size of 128. For sampled softmax loss, we sample 128 negatives for denser datasets like ML-1M, ML-20M, and Amazon Books, and 512 negatives otherwise. Given that there exists only one group of features in the sequential setting, we use learned MLPs to project user embedding at step $t$ to $k_u$ embeddings, and item embedding to $k_x$ embeddings for MoL and DeepFM. For other parameters, we follow reported optimal settings when those exist, and grid-searched parameters otherwise. Detailed configurations are in Appendix~\ref{sec:appendix-hparam-settings}. All models are implemented in PyTorch and are run on a single A100 80GB GPU.



Experiment results can be found in Table~\ref{tab:public_movielens} and Table~\ref{tab:public_amazon_reviews}. \textit{Dot product + SS} significantly outperforms \textit{baseline (BCE)} in all settings, confirming the importance of sampled softmax loss in retrieval setting. 
Non-dot product based methods (e.g., NeuMF and MoL) tend to significantly outperform dot products on datasets that are not extremely sparse (ML-1M, ML-20M, and Amazon Books). Dot products remain competitive for highly sparse datasets, e.g., Beauty and Games (avg $<10$ interactions per item). 

\begin{table*}
  \caption{MovieLens (-1M and -20M) results on top of sequential methods.}
  \label{tab:public_movielens}
  \vspace{-1em}
  \begin{tabular}{clllllll}
    \toprule
                        & method       & hr@1            & \underline{hr@10}             & hr@50           & hr@500          & mrr               \\ 
       \midrule
\multirow{5}{*}{ML-1M} & baseline (BCE)  & 0.0409           & \underline{0.2230}           & 0.4929           & 0.8594          & 0.1018      \\ 
                       & Dot product + SS  & 0.0669 (+63.6\%) & \underline{0.2821 (+26.5\%)} & 0.5435 (+10.3\%) & 0.8618 (+0.3\%) & 0.1363 (+33.9\%)   \\ 
                       & MLP + SS      & 0.0576 (+40.9\%) & \underline{0.2805 (+25.8\%)} & 0.5419 (+9.9\%)  & 0.8551 (-0.5\%) & 0.1290 (+26.8\%)  \\ 
                       & NeuMF + SS   & 0.0545 (+33.2\%) & \underline{0.2854 (+28.0\%)} & 0.5465 (+10.9\%) & 0.8672 (+0.9\%) & 0.1264 (+24.2\%)  \\ 
                       & DeepFM (8$\times$8) + SS  & 0.0626 (+53.0\%) & \underline{0.2856 (+28.1\%)} & 0.5526 (+12.1\%) & 0.8644 (+0.6\%) & 0.1335 (+31.2\%) \\
                       & MoL (8$\times$8) + SS & \textbf{0.0675 (+65.2\%)} & \underline{\textbf{0.3079 (+38.1\%)}} & \textbf{0.5657 (+14.8\%)} & \textbf{0.8677 (+1.0\%)} & \textbf{0.1433 (+40.8\%)} \\ 
       \midrule
\multirow{5}{*}{ML-20M} & baseline (BCE)   & 0.0275            & \underline{0.1756}           & 0.4467            & 0.8543          & 0.0782 \\
                        & Dot product + SS  & 0.0620 (+125.1\%) & \underline{0.2862 (+63.0\%)} & 0.5488 (+22.9\%)  & 0.8721 (+2.1\%) & 0.1344 (+71.9\%) \\ 
                        & MLP + SS    & 0.0687 (+149.4\%) & \underline{0.2936 (+67.2\%)} & 0.5556 (+24.4\%)  & 0.8756 (+2.5\%) & 0.1416 (+81.1\%) \\ 
                        & NeuMF + SS   & 0.0645 (+134.1\%) & \underline{0.2911 (+65.7\%)} & 0.5548 (+24.2\%)  & 0.8760 (+2.5\%) & 0.1376 (+76.0\%) \\ 
                        & DeepFM (8$\times$8) + SS  & 0.0657 (+138.7\%) & \underline{0.2932 (+67.0\%)} & 0.5542 (+24.1\%)  & 0.8763 (+2.6\%) & 0.1394 (+78.2\%) \\
                        & MoL (8$\times$8) + SS    & \textbf{0.0758 (+175.3\%)} & \underline{\textbf{0.3114 (+77.3\%)}} & \textbf{0.5727 (+28.2\%)} & \textbf{0.8847 (+3.6\%)} & \textbf{0.1528 (+95.3\%)} \\
  \bottomrule
\end{tabular}
\end{table*}

\begin{table}[b]
  \vspace{-1em}
  \caption{Rank of learned $\phi(u, x)$ on ML-1M ($d = 50$). Note that a high rank matrix does not always imply high hit rate.}
  \vspace{-0.9em}
  \label{tab:public_ml_1m_rank}
  \begin{tabular}{crrrrr}
    \toprule
                          & dot product & MLP   & NeuMF & DeepFM & MoL \\
    \midrule 
    Rank                  &  50         & 227   & 386   & 304    & 1864 \\
    \bottomrule
  \end{tabular}
\end{table}

\begin{table*}
  \caption{Amazon Reviews (Beauty, Games, and Books) results on top of sequential methods.}
  \vspace{-0.9em}
  \label{tab:public_amazon_reviews}
  \begin{tabular}{clllllll}
    \toprule
                        & method       & hr@10            & hr@50            & \underline{hr@200}            & hr@500                  & mrr                    \\ 
       \midrule
\multirow{5}{*}{Beauty} & baseline (BCE)  & 0.0256            & 0.0743           & \underline{0.1604}           & 0.2370               & 0.0126 \\
                        & Dot product + SS & \textbf{0.0655 (+155.4\%)} & \textbf{0.1379 (+85.5\%)} & \underline{\textbf{0.2271 (+41.6\%)}}   & 0.2914 (+23.0\%) & \textbf{0.0333 (+164.4\%)}  \\ 
                        & MLP + SS          & 0.0393 (+53.4\%)  & 0.1005 (+35.2\%) & \underline{0.1864 (+16.2\%)} & 0.2465 (+4.0\%)           & 0.0189 (+49.8\%) \\
                        & NeuMF + SS      & 0.0535 (+108.9\%) & 0.1247 (+67.8\%) & \underline{0.2152 (+34.2\%)} & 0.2818 (+18.9\%)          & 0.0264 (+109.6\%)                              \\ 
                        & DeepFM (4$\times$4) + SS & 0.0494 (+92.9\%)  & 0.1271 (+71.0\%) & \underline{0.2238 (+39.5\%)} & 0.2953 (+24.6\%)          & 0.0241 (+91.3\%) \\
                        & MoL (4$\times$4) + SS  & 0.0541 (+111.0\%) & 0.1288 (+73.3\%) & \underline{0.2246 (+40.0\%)} & \textbf{0.2961 (+25.0\%)} & 0.0262 (+107.5\%)  \\
       \midrule
\multirow{5}{*}{Games}  & baseline (BCE) & 0.0794                     & 0.2209           & \underline{0.4217}          & 0.5642          & 0.0383                          \\  
                        & Dot product + SS    & \textbf{0.1246 (+57.0\%)} & 0.2771 (+25.4\%) & \underline{0.4505 (+6.8\%)} & 0.5736 (+1.7\%) & \textbf{0.0600 (+56.7\%)}      \\  
                        & MLP + SS           & 0.1066 (+34.3\%)            & 0.2537 (+14.9\%) & \underline{0.4293 (+1.8\%)} & 0.5582 (-1.1\%) & 0.0498 (+30.1\%)                \\  
                        & NeuMF + SS        & 0.1175 (+48.0\%)          & 0.2722 (+23.2\%) & \underline{0.4479 (+6.2\%)} & 0.5755 (+2.0\%) & 0.0559 (+46.0\%)                 \\  
                        & DeepFM (4$\times$4) + SS    & 0.1216 (+53.2\%)            & 0.2817 (+27.5\%) & \underline{0.4690 (+11.2\%)} & \textbf{0.5965 (+5.7\%)} & 0.0576 (+50.5\%) \\
                        & MoL (4$\times$4) + SS    & 0.1221 (+53.9\%) & \textbf{0.2834 (+28.3\%)} & \underline{\textbf{0.4713 (+11.8\%)}} & \textbf{0.5964 (+5.7\%)} & 0.0586 (+53.2\%) \\

       \midrule
\multirow{5}{*}{Books}  & baseline (BCE)   & 0.0247            & 0.0660           & \underline{0.1370}    &  0.2079       & 0.0123 \\
                        & Dot product + SS & 0.0317 (+28.3\%)  & 0.0814 (+23.3\%) & \underline{0.1588 (+15.9\%)}  & 0.2295 (+10.4\%) & 0.0156 (+26.8\%) \\
                        & MLP + SS         & 0.0297 (+20.2\%)  & 0.0759 (+15.0\%) & \underline{0.1500 (+9.5\%)}   & 0.2190 (+5.3\%)  & 0.0144 (+17.1\%) \\
                        & NeuMF + SS       & 0.0358 (+45.0\%)  & 0.0871 (+32.0\%) & \underline{0.1644 (+20.0\%)}  & 0.2338 (+12.5\%) & 0.0178 (+44.7\%) \\
                        & DeepFM (8$\times$8) + SS & 0.0361 (+46.2\%)  & 0.0905 (+37.1\%) & \underline{0.1706 (+24.5\%)}  & 0.2414 (+16.1\%) & 0.0179 (+45.5\%) \\
                        & MoL (8$\times$8) + SS   & \textbf{0.0388 (+57.1\%)}  & \textbf{0.0934 (+41.5\%)} & \underline{\textbf{0.1751 (+27.8\%)}}  & \textbf{0.2479 (+19.2\%)} & \textbf{0.0194 (+57.7\%)} \\
  \bottomrule
\end{tabular}
\end{table*}

We also observe that we can indeed obtain high rank $\phi(u, x)$ distributions with non-dot-product methods, as shown in Table~\ref{tab:public_ml_1m_rank}. 


\subsubsection{Ablation Studies}
\label{sec:exp-public-ds-ablation}

We run the following ablation studies to understand how different components in MoL contribute to overall performance. We fix the baseline as the default MoL configuration, which differs from Table~\ref{tab:public_movielens} and~\ref{tab:public_amazon_reviews} in that we enabled component-level hypersphere embeddings for Amazon Beauty and Games datasets. 
\begin{itemize}[leftmargin=*]
    \item \textit{no-l2-norm}: Ablate component-level hypersphere embeddings. 
    \item \textit{no-gating-dropout}: Ablate dropout in gating. 
    \item \textit{50\%-$k_u \times k_x$}: Use 50\% fewer mixture components for $k_u$ and $k_x$.
    \item \textit{25\%-negatives}: Use 75\% fewer negative examples. 
\end{itemize}
Results can be found in Table~\ref{tbl:ablation_studies_seq_methods}. Most components show consistent gains except on the sparse Amazon datasets. Those that consistently contribute to gains ($> 2\%$ are underlined) include component-level hypersphere embeddings, gating dropout, and number of negatives. 

\begin{table*}
  \caption{Ablation studies on top of MoL and sequential methods.}
  \vspace{-0.9em}
  \label{tbl:ablation_studies_seq_methods}
  \begin{tabular}{clllll}
    \toprule
                          & ML-1M (HR@10)       & ML-20M (HR@10)     & Beauty (HR@200)    & Games (HR@200)     & Books (HR@200) \\
       \midrule
       MoL                & 0.3079               & 0.3114              & 0.2088             & 0.4618             & 0.1751\\
       no-l2-norm         & \underline{0.2894 (-6.0\%)}    & \underline{0.3032 (-2.6\%)}     & \textit{0.2246 (+7.5\%)}    & \textit{0.4713 (+2.0\%)}    & \underline{0.1695 (-3.2\%)} \\
   no-gating-dropout & \underline{0.2964 (-3.8\%)}    & 0.3100 (-0.5\%)     & \textit{0.2148 (+2.8\%)}    & 0.4592 (-0.6\%)    & \underline{0.1674 (-4.4\%)} \\
       50\%-$k_u\times k_x$   & 0.3041 (-1.2\%)    & \underline{0.3022 (-2.9\%)}     & 0.2076 (-0.6\%)    & 0.2076 (-0.6\%)   & 0.1719 (-1.8\%) \\
       25\%-negatives     & \underline{0.2972 (-3.5\%)}    & \underline{0.2907 (-6.7\%)}  & \underline{0.2046 (-2.0\%)}   & \textit{0.4627 (+0.2\%)}    & \textit{0.1808 (+3.3\%)} \\
    \bottomrule
    \end{tabular}
\end{table*}





\subsection{Industrial Datasets}
We next consider a non-sequential model setup, and evaluate model performance on a proprietary industrial dataset against strong baselines. All models share the same features, labels, and losses (incl. number of negatives), and all non-MoL setups use similar compute as the most expensive MoL setup (32$\times$4). Equation~\ref{eq:adaptive-logit-selection-user-side} was used to compress user-side and item-side embeddings for DeepFM~\cite{deepfm_ijcai17}.



Results are shown in Table~\ref{tbl:industry_eval_result}. With sufficient model capacity, 
our proposed MoL model outperforms all baselines under the same compute budget. The best MoL model improves HR@50 by 23.3\% compared with Dot product and 5.0\% compared with DeepFM.

\subsubsection{Hierarchical Retrieval}
\label{sec:exp-hierarchical-retrieval-eval}

\begin{figure}[t]
  \centering
  \includegraphics[width=0.85\linewidth]{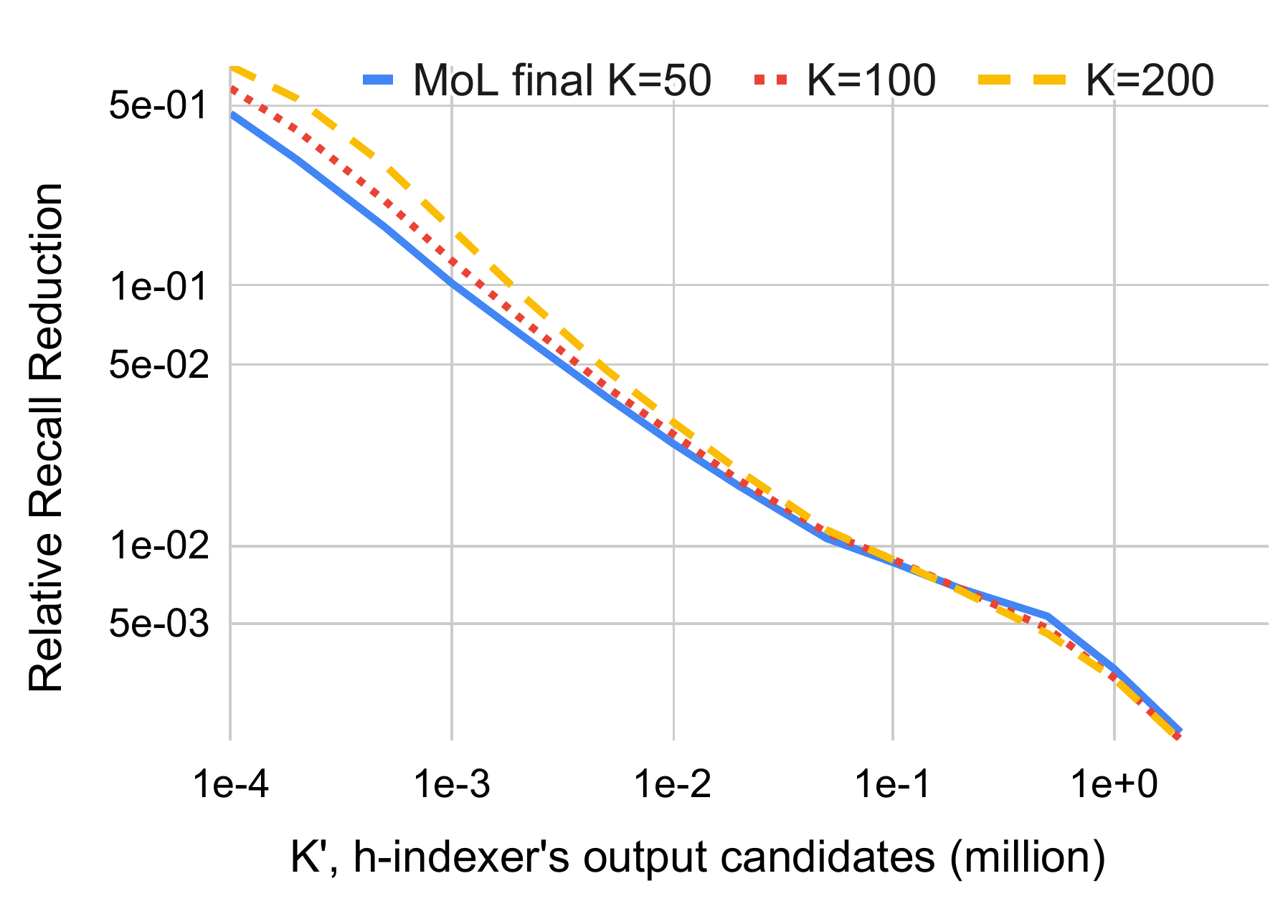}
  \includegraphics[width=0.85\linewidth]{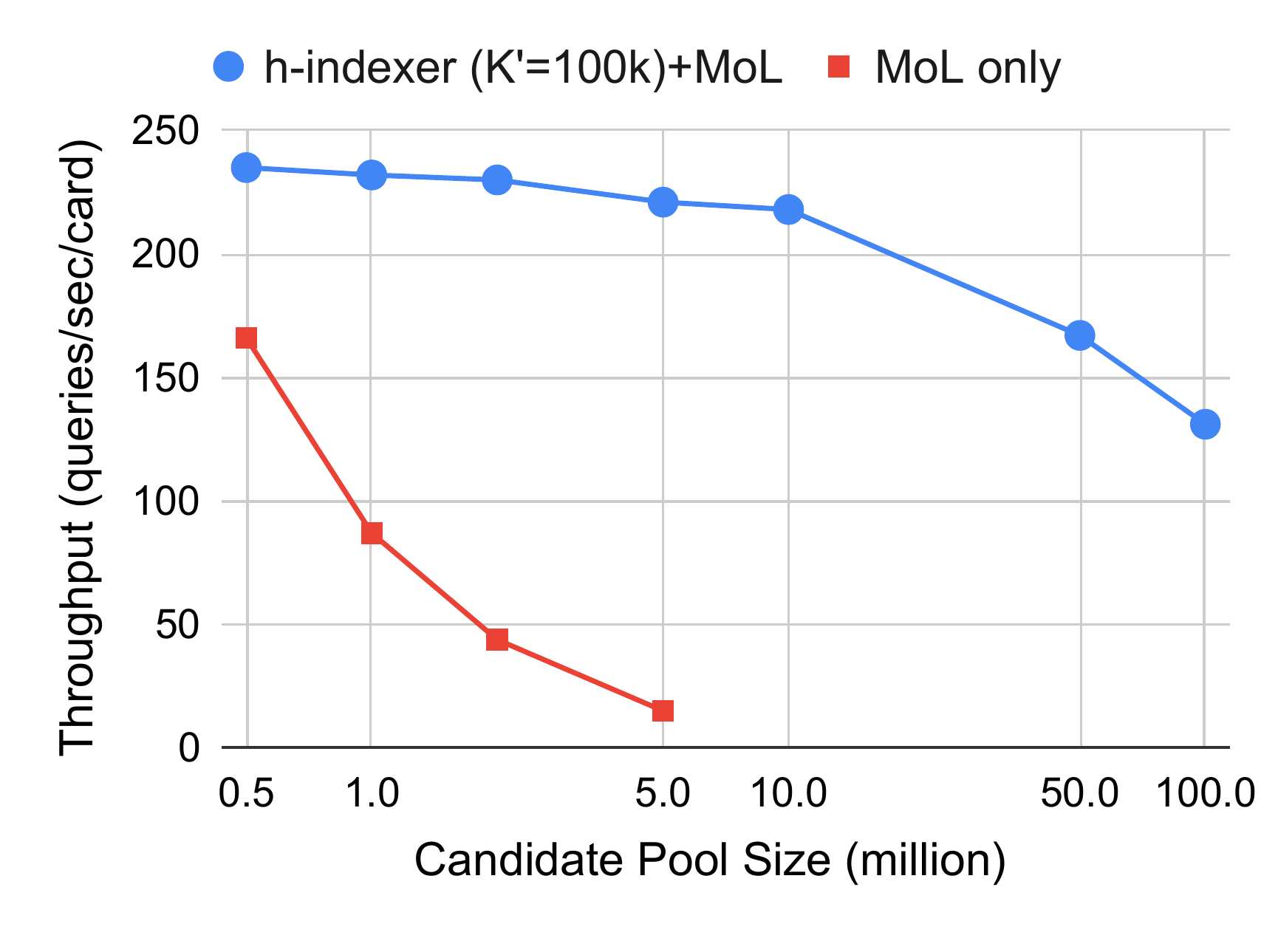}
  \vspace{-2em}
  \caption{Recall and serving cost of the two-stage h-indexer/MoL model: (a) relative recall ratio vs MoL-only model under varying $k'$ over 10M items, and (b) throughput of two-stage vs one-stage over different pool sizes.} 
  \label{fig:prefilter_k_prime_vs_recall}
\end{figure}

We next evaluate how our hierarchical retrieval design affects model quality and scalability. Figure~\ref{fig:prefilter_k_prime_vs_recall} compares the performance of MoL with and without h-indexer. We observe that h-indexer does not degrade recall for suitably chosen values of $K'$ (e.g., $10^5$). While MoL + h-indexer with $K' = 10^5$ has almost the same recall as the MoL-only model ($K'=X$), its throughput is less sensitive to corpus size $X$ and enables us to scale up to 100M items on a single GPU.

\subsubsection{Effect of Different Mixture Components.}
We analyze the role of mixture components to understand how model performance scales with model capacity in Table~\ref{tbl:industry_eval_result}. As we increase number of mixture components from 8$\times$4 to 32$\times$4, we can observe significant performance improvement in each scale-up. Meanwhile, the performance of Dot product starts to saturate at dim=256. 

\subsubsection{Popularity Bias.} 
\begin{figure}[t]
  \centering
  \vspace{-1em}
  \includegraphics[width=\linewidth]{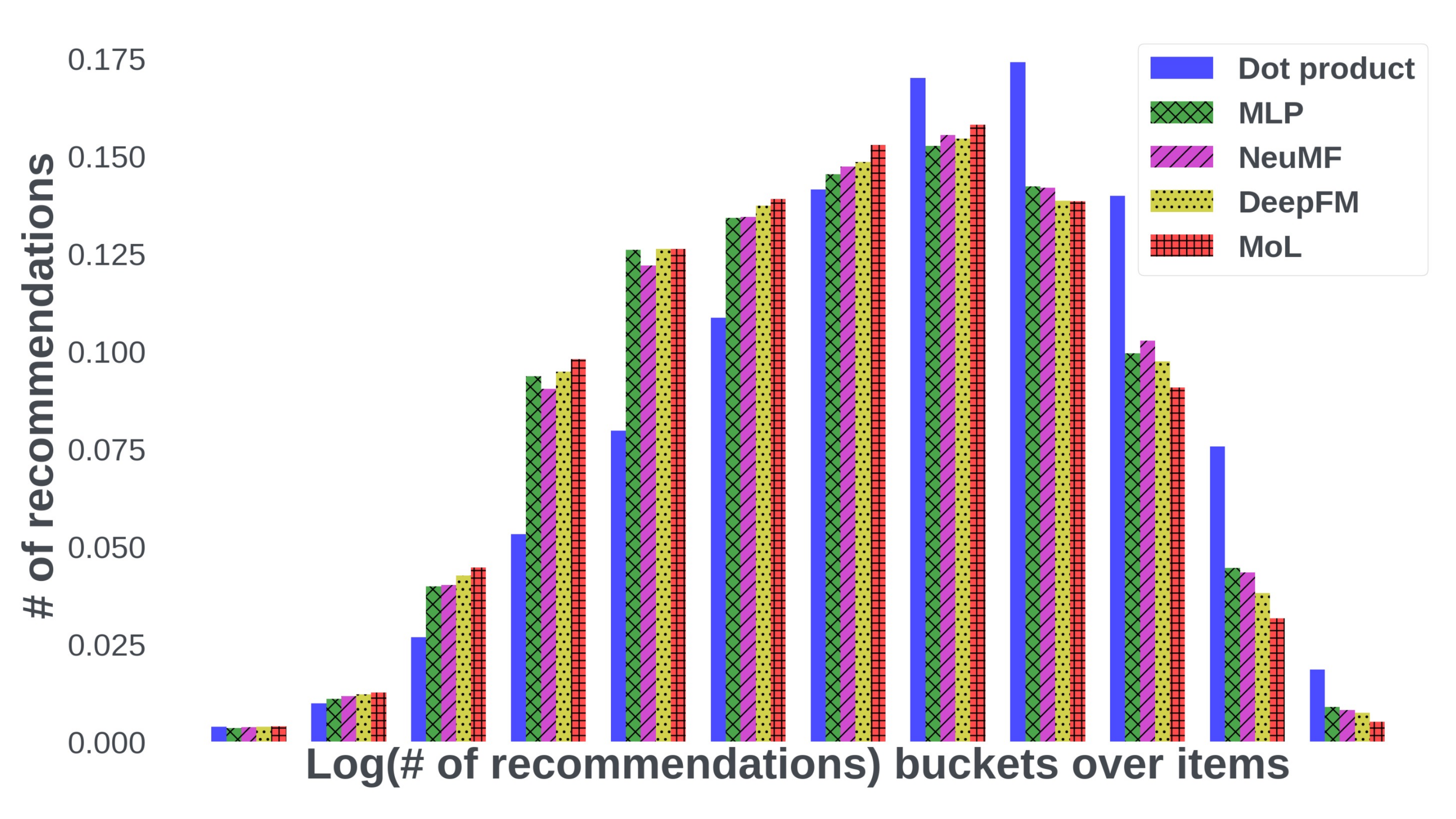}
  \vspace{-2em}
  \caption{Distributions of recommendations over log-scaled recommendation frequency buckets. Lower bars in higher buckets indicates that fewer head items are shown.}
  \label{fig:popularity-bias}
  \vspace{-1.5em}
\end{figure}

We observe that MoL and other non-dot product methods also reduce popularity bias. 
We plot a histogram to show the distributions of recommended items in Figure~\ref{fig:popularity-bias}. Items recommended by MLP, NeuMF, DeepFM, and MoL have more coverage in tail segments and are less biased towards popular ones, 
with MoL showing the most significant reduction in head buckets. 

\begin{table}[b]
  \vspace{-1em}
  \caption{Offline results on a large-scale industrial dataset.}
  \vspace{-0.9em}
  \label{tbl:industry_eval_result}
  \begin{tabular}{cllll}
    \toprule
       method       & HR@10       & HR@50     & HR@100    & HR@200 \\
       \midrule
       Dot product (d=64) & 0.0843  &  0.2563   & 0.3712  & 0.5026 \\
       Dot product (d=128) & 0.0882  &  0.2676   & 0.3856  & 0.5204 \\
       Dot product (d=256) & 0.0916  &  0.2761   & 0.3954  & 0.5305 \\
       Dot product (d=512) & 0.0925  &  0.2754   & 0.3949  & 0.5299 \\
       \midrule
       MLP         & 0.0938   &  0.2656      & 0.3770     & 0.5054  \\
       \midrule
       NeuMF       & 0.1087   &  0.2976    & 0.4148  & 0.5479   \\
       \midrule
       DeepFM      & 0.1199  &  0.3242  & 0.4465 & 0.5810   \\
       \midrule
       MoL (16$\times$2)  & 0.1212  &  0.3221  & 0.4436 & 0.5768   \\
       MoL (8$\times$4)   & 0.1221  &  0.3232  & 0.4440 & 0.5763   \\
       MoL (16$\times$4)  & 0.1250  &  0.3283  & 0.4502 & 0.5834   \\
       MoL (32$\times$4)  & \textbf{0.1304}   &  \textbf{0.3404}  & \textbf{0.4628}  & \textbf{0.5961}  \\
    \bottomrule
    \end{tabular}
\end{table}



\vspace{-.3em}
\subsection{Online Experiments}
We conducted A/B testing for the proposed setup on a major recommendation surface at Meta serving billions of users. 
Over a period of 7 days, we observed a significant gain of $+3.2\%$ in our main topline metric and $+0.22\%$ completion rate 
compared with baseline model. 


We also compared the latency and throughput between the older production CPU model and the new MoL + h-indexer model on GPUs. The median serving GPU latency of our proposed architecture is 30ms, comparable with optimized CPU MIPS baseline (20ms).  
The latency regression is a result of request batching optimization, where we let the service wait for up to 10 ms to construct a batch from live requests. 
Despite high FLOPs, our design does not significantly increase serving cost due to its high arithmetic intensity. A comparable GPU MIPS implementation has 13ms latency (before batching) and 1.8$\times$ throughput vs MoL + h-indexer. 




\section{Related Work}
\label{sec:related-work}

\indent\indent
\textit{Dot Product as an Embedding Combiner.} 
Dot products have been widely used to combine model outputs with class information in seminal work across image classifications (e.g., AlexNet~\cite{alexnet_neurips12}), natural language processing (e.g., Transformers~\cite{transformers_goog_neurips17} and BERT~\cite{bert_acl19}), and information retrieval (e.g., YouTube DNNs~\cite{ytdnn_goog_recsys16}).

Recently, softmax layer is observed to be a bottleneck 
for some NLP applications~\cite{mos_iclr18}, and averaging multiple softmaxes has been found to be helpful in improving model perplexity. Averaging softmaxes, however, requires computation of normalization factor 
for each softmax, which is costly for NLP vocabularies in the range of $10^4-10^5$~\cite{mos_iclr18} and cost-prohibitive for modern IR settings that need to deal with millions to billions of items~\cite{alibaba_scale_kdd18,pixie_pins_www18}. Our algorithm not only reduces computational cost, but also enables better efficiency via embedding sharing as discussed in Section~\ref{sec:alg-emb-sharing-adaptive-logits}.


\textit{Second- and Higher-Order Interaction Modeling.} Motivated by observations such as people often downloading food delivery apps at mealtime~\cite{deepfm_ijcai17}, feature interactions have been recognized as an important factor for achieving good recommendation performance in various works, including Factorization Machines (FMs)~\cite{fm_rendle_icdm10} and their deep neural network variants~\cite{deepfm_ijcai17,afm_ijcai17}, product-based NNs~\cite{pnn_icdm16},  DCNs~\cite{dcn_adkdd17}, and so on. The applications of these approaches have been limited to CTR predictions, or ranking. We believe that this is largely due to the difficulty of generalizing MLPs to long tail as discussed by Rendle et al. in~\cite{ncf_mf_goog_recsys20}. Compared with Attention FMs~\cite{afm_ijcai17} and PNNs~\cite{pnn_icdm16}, MoL leverages adaptive embedding compression and gating decomposition to significantly reduce compute and memory usage, and l2 norm'ed features to improve generalization. 


\textit{Multi-Interest Recommendations and Short-/Long-Term Interests.} A line of work related to interaction modeling is multi-interest embeddings~\cite{mind_baba_cikm19,sine_baba_wsdm21,sdm_baba_cikm19} which aims to capture diverse preferences of a user over different topics and time periods.
For cases where the embedding combiner is not user- \textit{and} item- dependent, like SINE and SDM~\cite{sine_baba_wsdm21,sdm_baba_cikm19}, the resulting model could be more expressive but still necessarily low rank as we've shown in Equation~\ref{eq:mol_low_rank_variant}. Item-dependent approaches like MIND~\cite{mind_baba_cikm19} can be viewed as a special form of MoL where there is exactly one item-side embedding and gating function takes a specific form (e.g., dot-product attention).


\textit{Learned Discrete Structures for Retrieval.} Yet another approach is to learn explicit structures so that retrieval can be viewed as a root-to-leaf traversal, beam search, or operating on top of binary codes directly~\cite{tdm_baba_kdd18, jtm_baba_neurips19, otm_baba_icml20, bccg_cikm19, dr_cikm21}. In particular, TDM~\cite{tdm_baba_kdd18} first enabled expressive neural networks to be applied at retrieval time. These approaches are more sensitive to hyperparameters, and if heavy branching is involved, can be hard to scale on accelerators. 


\textit{Accelerating Inference for Retrieval.}  In the traditional dot product retrieval setting, or MIPS (Maximum Inner Product Search), product quantization and hashing techniques~\cite{alsh_ping_neurips2014,quant_mips_kumar_aistats16} have been well studied and are widely used in the non-accelerator retrieval settings. 

Partitioning the item space is a complementary approach to speed up inference by reducing search space~\cite{clusteringss_tkde02,hdss_sigmod11,conetree_mips_kdd12,metrictree_mips_cikm12,tdm_baba_kdd18,jtm_baba_neurips19,dr_cikm21}. Search space can be partitioned with clustering~\cite{clusteringss_tkde02,hdss_sigmod11}, spatial-partitioning trees~\cite{conetree_mips_kdd12,metrictree_mips_cikm12}, or with approaches where the partition strategy is learned~\cite{tdm_baba_kdd18,jtm_baba_neurips19,dr_cikm21}. This line of work can be viewed as alternatives to \textit{h-indexer} for hierarchical retrieval settings. 

A recent line of work investigates efficient MIPS-based KNN retrieval on accelerators~\cite{faiss_tbd21, tpuknn_goog_neurips22}. There have not been significant work on non-MIPS setups on accelerators to date. 
\vspace{-0.5em}

\section{Conclusions and Future Work}
\label{sec:conclusions}

We have presented new algorithms, including \textit{mixture of logits} and \textit{h-indexer} that enable non-dot-product retrieval models to run efficiently on accelerators with latency and throughput comparable to MIPS retrieval setups. These algorithms, which are compatible with various retrieval setups, significantly outperform baselines on both public and large industrial datasets with better recommendation quality and reduced popularity bias.

Our work lays the foundation for many interesting future work for non-dot-product based neural retrieval models. For instance, one direction is to make the mixture components in MoL arbitrarily complex neural networks instead of dot products. Another direction could be to make a cascading retrieval setup end-to-end learnable.

\section{Acknowledgements}
We'd like to thank Yinghai Lu, Andrew Tulloch, Pawel Garbacki, Jongsoo Park, Jeff Johnson, Dmytro Ivenchko, Yi Wang, Rui Zhang, Xianjie Chen for various discussions, Jie Hua, Michael He, Pengchao Wang, Chao Yang, Jiacheng Feng, Rui Jian, Hao Lin, Brandon Burroughs, Shripad Thite, Eric Xu, Qing Liu, Jeevan Gyawali, Lei Chen, Bi Xue, Mengchi Zhang, Jianyu Huang, Haixin Liu, Jiecao Yu, Sarunya Pumma, Dhruv Choudhary, Cheng Cheng, Zhiyun Zhang, Will Feng, Fanny Yang, Hao Lu, Wenlei Xie, Bin Wen, Emily Sun, Tianshu Bao, Zexi Mao, Bugra Akyildiz, Hui Zhang, Xinyao Hu, Yun Mao, Kaushik Veeraraghavan for infra/product support, and Shilin Ding, Lin Qiao, Hong Yan, Andre Rohe, Lars Backstrom for overall support of the project.

\bibliographystyle{ACM-Reference-Format}
\bibliography{sample-base}

\appendix

\section{Hyperparameter settings (public datasets)}
\label{sec:appendix-hparam-settings}


For ML-1M, Amazon Beauty, and Amazon Games datasets, we follow the original optimal hyperparams for SASRec reported in~\cite{sasrec_icdm18}, specifically two encoder layers ($b=2$) with one attention head ($h=1$), and a dropout rate of 0.2 for MovieLens and 0.5 for Amazon Reviews datasets. For ML-20M and Amazon Books, we kept dropout rate identical, grid-searched embedding dimension, number of encoder layers and attention heads for SASRec, and obtained $d=256$, $b=4$, $h=8$ for ML-20M and $d=64$, $b=4$, $h=4$ for Books.

We grid searched MoL parameters ($k_u$, $k_x$) in $(2, 2)$, $(4, 4)$, $(8, 8)$ and fixed component-level embedding dimension to $32$. For DeepFM experiments, we used the exact same configuration for user- and item- side projections as MoL, and tuned hidden dim and dropout rate additionally. Given the high computational costs of running experiments on ML-20M and Amazon Books datasets (up to 3d for 300 epochs), we shared NeuMF/DeepFM/MoL hparams between those two groups and ML-1M. Detailed configurations can be found in Table~\ref{tab:deprecated_ml_20m}.

\begin{table*}[htb]
  \caption{Hyperparameters used for public datasets}
  \vspace{-1em}
  \label{tab:deprecated_ml_20m}
  \begin{tabular}{llrrrrr}
    \toprule
                                 &                      & ML-1M         & ML-20M         & Beauty      & Games      & Books    \\
    \midrule
    \multirow{4}{*}{SASRec}      & embedding dim $d$    & 50            & 256           & 50          & 50          & 64      \\
                                 & encoder blocks $b$   & 2             & 4             & 2           & 2           & 4       \\
                                 & attention heads $h$  & 1             & 8             & 1           & 1           & 4       \\
                                 & dropout rate         & 0.2           & 0.2           & 0.5         & 0.5         & 0.5     \\
    \midrule
    \multirow{1}{*}{Dot Product} & temperature          & 20            & 20           & 20          & 20           & 20   \\
    \midrule
    \multirow{3}{*}{MLP}         & hidden layer size              & 512           & 512           & 128          & 128    & 256 \\
                                 & dropout rate                   & 0.1           & 0.1           & 0.2          & 0.2    & 0.2 \\
                                 & Inference FLOPs per ($x$, $u$) & 50.5K         & 256.5K        & 12.6K        & 12.6K  & 32.3K\\
    \midrule
    \multirow{6}{*}{NeuMF}       & GMF dim                        & 32            & 32            & 32           & 32     & 32 \\
                                 & MLP hidden dim                 & 256           & 256           & 128          & 128    & 256 \\
                                 & MLP output dim                 & 64            & 64            & 64           & 64     & 64 \\
                                 & dropout rate                   & 0.1           & 0.1           & 0.2          & 0.2    & 0.1 \\
                                 & final MLP hidden dim           & 256           & 256           & 128          & 128    & 256 \\
                                 & Inference FLOPs per ($x$, $u$)  & 49.3K         & 152.3K        & 24.6K        & 24.6K  & 56.3K \\
    \midrule
    \multirow{3}{*}{FM}          & hidden dim for output layer    & 256           & 256           & 128          & 128    & 256 \\
                                 & prediction layer dropout rate  & 0.2           & 0.2           & 0.3          & 0.3    & 0.2  \\
                                 & Inference FLOPs per ($x$, $u$)  & 197.6K        & 405K          & 10.7K        & 10.7K  & 203.6K \\
    \midrule
    \multirow{4}{*}{MoL}         & ($k_u$ x $k_x$ x embedding dim)             & (8 x 8 x 32)  & (8 x 8 x 32) & (4 x 4 x 32) & (4 x 4 x 32) & (8 x 8 x 32) \\
                                 & hidden dim for gating MLPs                  & 128           & 128          & 128          & 128          & 128          \\
                                 & hidden dim for embedding proj. MLPs         & 512           & 512          & N/A          & N/A          & 512          \\
                                 & item-side embedding proj. MLP dropout rate  & 0.1           & 0.1          & 0.3          & 0.3          & 0.1          \\
                                 & gating softmax dropout rate                 & 0.2           & 0.2          & -            & -            & 0.2           \\
                                 & gating input dropout rate                   & -             & -            & 0.2          & 0.2          & -             \\
                                 & component-level hypersphere embeddings      & On            & On           & Off          & Off          &  On \\
                                 & $\tau$                                      & 20            & 20           & 20           & 20           & 20 \\
                                 & Inference FLOPs per ($x$, $u$)               & 211.7K        & 445.3K       & 12.9K        & 12.9K        & 227.6K\\
                                 & Non-cachable inference FLOPs per ($x$, $u$)  & 32.9K         & 33.1K        & 4.4K          & 4.4K        & 21.5K \\
  \bottomrule
\end{tabular}
\end{table*}

\end{document}